\documentclass[letterpaper]{article} 
\usepackage{aaai2026}  
\usepackage{times}  
\usepackage{helvet}  
\usepackage{courier}  
\usepackage[hyphens]{url}  
\usepackage{graphicx} 
\usepackage{natbib}  
\usepackage{caption} 
\usepackage{algorithm}
\usepackage{algorithmic}
\usepackage{amssymb}
\usepackage{multirow}
\usepackage{booktabs}

\usepackage{newfloat}
\usepackage{listings}
\DeclareCaptionStyle{ruled}{labelfont=normalfont,labelsep=colon,strut=off} 
\floatstyle{ruled}
\newfloat{listing}{tb}{lst}{}
\floatname{listing}{Listing}

\newcommand{\figref}[1]{Fig.~\ref{#1}}
\newcommand{\tabref}[1]{Tab.~\ref{#1}}

\usepackage[table]{xcolor}
\definecolor{colorfirst}{RGB}{255, 204, 204}
\definecolor{colorsecond}{RGB}{255, 230, 204}
\definecolor{colorthird}{RGB}{255, 251, 214}

\newcommand{\ie}{\emph{i.e}. }

\newcommand{\etc}{\emph{etc}. }
\let\oldtwocolumn\twocolumn

\begin{document}


\title{FashionMAC: Deformation-Free Fashion Image Generation  \\
with Fine-Grained Model Appearance Customization}

\author {
    Rong Zhang\textsuperscript{\rm 1},
    Jinxiao Li\textsuperscript{\rm 1},
    Jingnan Wang\textsuperscript{\rm 1},
    Zhiwen Zuo\textsuperscript{\rm 1}\footnotemark[1], \\
    Jianfeng Dong\textsuperscript{\rm 1},
    Wei Li\textsuperscript{\rm 2},
    Chi Wang\textsuperscript{\rm 3},
    Weiwei Xu\textsuperscript{\rm 3},
    Xun Wang\textsuperscript{\rm 1}\thanks{Corresponding author.}
}

\affiliations {
    \textsuperscript{\rm 1}Zhejiang Gongshang University\\
    \textsuperscript{\rm 2}Nanjing University \\
    \textsuperscript{\rm 3}Zhejiang University\\
    zhangrong@zjgsu.edu.cn, \{module8627, wangjingnan751\}@gmail.com, zzw@zjgsu.edu.cn, \\
    \{dongjf24, liweimcc\}@gmail.com, wangchi1995@zju.edu.cn, xww@cad.zju.edu.cn, wx@zjgsu.edu.cn 
}

\maketitle



\begin{abstract}
Garment-centric fashion image generation aims to synthesize realistic and controllable human models dressing a given garment, which has attracted growing interest due to its practical applications in e-commerce. The key challenges of the task lie in two aspects: (1) faithfully preserving the garment details, and (2) gaining fine-grained controllability over the model's appearance. 
Existing methods typically require performing garment deformation in the generation process, which often leads to garment texture distortions. Also, they fail to control the fine-grained attributes of the generated models, due to the lack of specifically designed mechanisms.
To address these issues, we propose FashionMAC, a novel diffusion-based deformation-free framework that achieves high-quality and controllable fashion showcase image generation. The core idea of our framework is to eliminate the need for performing garment deformation and directly outpaint the garment segmented from a dressed person, which enables faithful preservation of the intricate garment details.
Moreover, we propose a novel region-adaptive decoupled attention (RADA) mechanism along with a chained mask injection strategy to achieve fine-grained appearance controllability over the synthesized human models.
Specifically, RADA adaptively predicts the generated regions for each fine-grained text attribute and enforces the text attribute to focus on the predicted regions by a chained mask injection strategy, significantly enhancing the visual fidelity and the controllability. Extensive experiments validate the superior performance of our framework compared to existing state-of-the-art methods. 
\end{abstract}

\begin{links}
    \link{Project}{https://github.com/module8627/FashionMAC}
    \link{Supp.}{https://arxiv.org/abs/2511.14031}
\end{links}

\section{Introduction}\label{sec:intro}

\begin{figure}[!t]
  \centering
  \includegraphics[width=\linewidth]{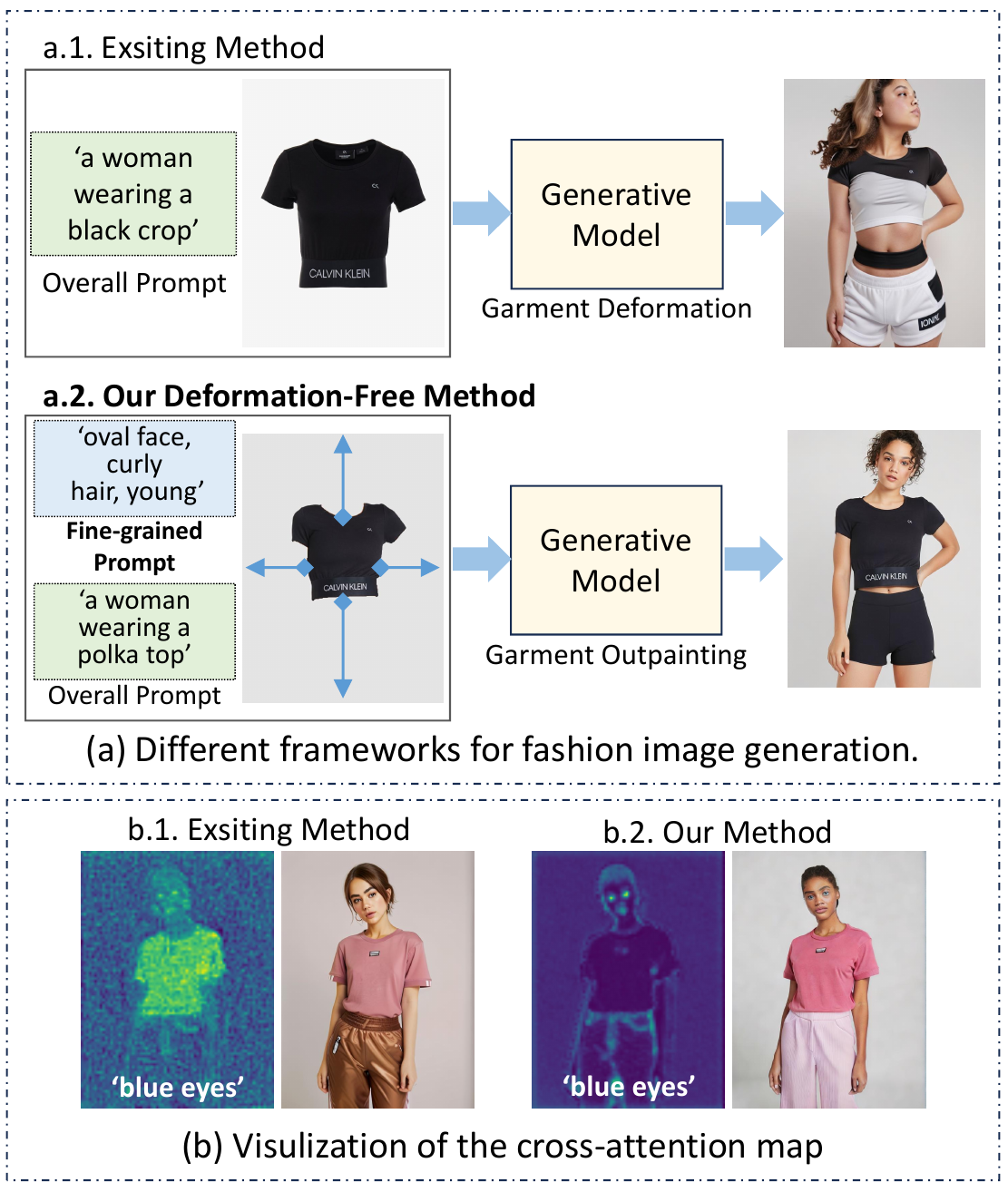}
   \caption{(a): Framework comparison of fashion image generation methods. (b): Visualization of the cross attention-map. Note that our method can preserve details better and enable accurate fine-grained text customization. }
   \label{fig:vs}
\end{figure}

The rapid advancement of AI technology is revolutionizing numerous industries including fashion e-commerce~\cite{dong2021fine, zhu2023tryondiffusion,yang2024texture, dong2025open}. Recently, a new task called garment-centric fashion image generation~\cite{chen2024magic, lin2025dreamfit, shen2025imagdressing} has gained increasing attention. Different from conventional fashion image synthesis tasks like virtual try-on, this line of research focuses on synthesizing realistic and controllable human models wearing a given garment (see Fig.~\ref{fig:vs} (a)), enabling automated fashion showcase creation for online fashion retailers. 
The key challenges of the task lie in two aspects: 1) faithful preservation of garment details, and 2) fine-grained controllability over the generated model’s appearance.

\begin{figure*}[!t]
  \centering
  \includegraphics[width=0.95\textwidth]{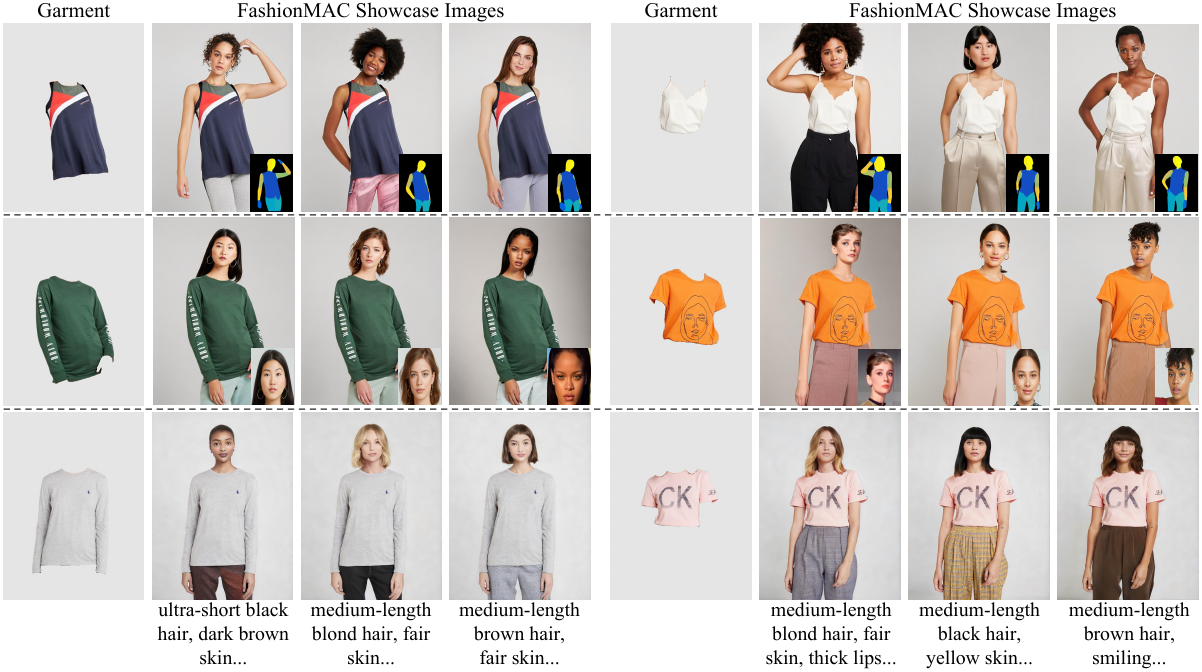}
   \caption{Given a garment image segmented from a dressed mannequin or a person, our method can generate fashion showcase images via garment-centric outpainting under the guidance of face images or fine-grained text attributes. From top to bottom: (1) The results conditioned on the automatically generated diverse pose maps. (2) The results conditioned on different face images. (3) The results conditioned on fine-grained text prompts. }
   \label{fig:teaser}
\end{figure*}

Many approaches have been proposed to tackle this task by leveraging the strong generative capability of pre-trained text-to-image diffusion models, \textit{e.g.}, latent diffusion models (LDMs)~\cite{LDM}. MagicClothing (MC)~\cite{chen2024magic} injects garment features into the LDM and conditions the generation process with textual descriptions to customize the model's appearance. Parts2Whole~\cite{huang2024parts} extends this paradigm by encoding hierarchical features from images of different human parts (\textit{e.g.}, face, hair, and clothes) and then integrating them into the generation network for region-aware synthesis.
DreamFit~\cite{lin2025dreamfit} incorporates a specifically tailored lightweight encoder for efficient generation.

Despite the advances, these methods fall short in preserving the intricate garment details and freely controlling the fine-grained appearance attributes of the synthesized human models.
On the one hand, the above methods typically involve deforming the reference garments to fit the synthesized model's poses in their generation process. However, as shown in ~\figref{fig:vs} (a), garment deformation may lead to texture distortions or detail losses in the generated images, which can seriously limit these methods' practical usage in real-world applications.
On the other hand, existing methods struggle to manipulate the fine-grained appearance attributes of the synthesized models (\textit{e.g.}, the eye color as shown in~\figref{fig:vs} (b)). 
As pointed out by previous literature~\cite{zhang2024enhancing}, the cross-attention layers in LDMs have a propensity to disproportionately focus on certain tokens while ignoring others during the generation process.
By examining the cross-attention maps of the fine-grained appearance text tokens in an existing method as shown in~\figref{fig:vs}(b), we observe that these tokens exhibit low activations toward their corresponding generation regions, while most of their energy is diverted to irrelevant areas. Therefore, we hypothesize that the problem stems from existing methods’ failure to attend to fine-grained attribute tokens during the generation process.

To address these limitations, we propose FashionMAC, a novel deformation-free garment-centric \textbf{Fashion} image generation framework
with fine-grained \textbf{M}odel \textbf{A}ppearance \textbf{C}ustomization. Instead of deforming the garment to fit a target pose, FashionMAC directly outpaints a dressed garment image to synthesize a realistic human model with the guidance of text descriptions or face images, as illustrated in \figref{fig:vs} (a). Such dressed garment images --- can be easily obtained by dressing a mannequin or having a person wear the clothes --- already contain natural deformations, allowing our method to bypass the complex deformation process. Our design avoids visual distortions caused by inaccurate warping and ensure faithful preservation of intricate garment details in the synthesized results. Thanks to the abundance of dressed-person images available online, our framework can be trained without requiring paired data of garments and corresponding human models, significantly improving data accessibility and real-world applicability.

Furthermore, we empower FashionMAC with the capability to control the fine-grained characteristics of the synthesized models by proposing a novel region-adaptive decoupled attention (RADA) mechanism. 
Different from existing methods that treat the input text prompt as a global one, we introduce RADA to process the overall text descriptions and fine-grained text attributes separately. 
Specifically, a fine-grained mask prediction module is proposed to adaptively localize the influence regions of the detailed text attributes and then these attributes are encouraged to attend to the predicted regions.
In addition, we further propose a chained mask injection (CMI) strategy that utilizes the masks predicted from the previous timestep to steer RADA in the current generation timestep, providing effective mask guidance for RADA in the denoising process. 
These carefully crafted designs enable the proposed framework to accurately align the influenced feature map regions with the fine-grained text attributes, significantly enhancing the proposed framework's controllability over the fine-grained text attributes.

To summarize, our key contributions are as follows:
\begin{itemize}
    \item We propose FashionMAC, a novel deformation-free garment-centric framework for fashion image generation, eliminating the need for garment deformation and ensuring faithful preservation of garment details.
    \item We introduce a novel region-adaptive decoupled attention mechanism along with a chained mask injection strategy to enable fine-grained model appearance customization. To the best of our knowledge, we are the first to investigate fine-grained model appearance customization in fashion image generation. 
    \item Extensive experiments demonstrate that our method outperforms existing approaches in terms of both visual quality and appearance controllability. 
\end{itemize}

\section{Related Works}

\subsection{Image-Based Virtual Try-on}
Image-based virtual try-on (VITON) task is designed to generate realistic try-on images based on the given garment image and the reference person image.
Most prior VITON methods consist of two main phases: the cloth warping phase and the try-on generation phase. In the first phase, thin plate spline (TPS)~\cite{duchon1977splines} and appearance flow~\cite{li2019dense} are commonly adopted for cloth warping. In the second phase, GANs~\cite{goodfellow2014generative} often play a pivotal role in refining try-on results. Recently, many researchers have developed methods based on LDMs~\cite{LDM} due to their superior image quality. Some approaches~\cite{gou2023taming,li2023virtual,wang2024fldm} still follow the previous two-phase framework, substituting GANs with LDMs in the second phase for synthesizing more realistic try-on results, while others~\cite{morelli2023ladi,zhu2023tryondiffusion,yang2024texture,choi2024improving,kim2024stableviton,zeng2024cat} build end-to-end frameworks by introducing garment encoders into LDMs and then utilizing attention modules to perform implicit garment warping. Nonetheless, few works have paid attention to apparel showcase image generation. Magic Clothing (MC)~\cite{chen2024magic} is the most related work to ours in this area, which customizes characters wearing the target garment with diverse text prompts. However, compared to our method, MC falls short in cloth details and lacks precise control over the fine-grained attributes of the generated model's appearance.

\begin{figure*}[t]
  \centering
  \includegraphics[width=0.9\linewidth]{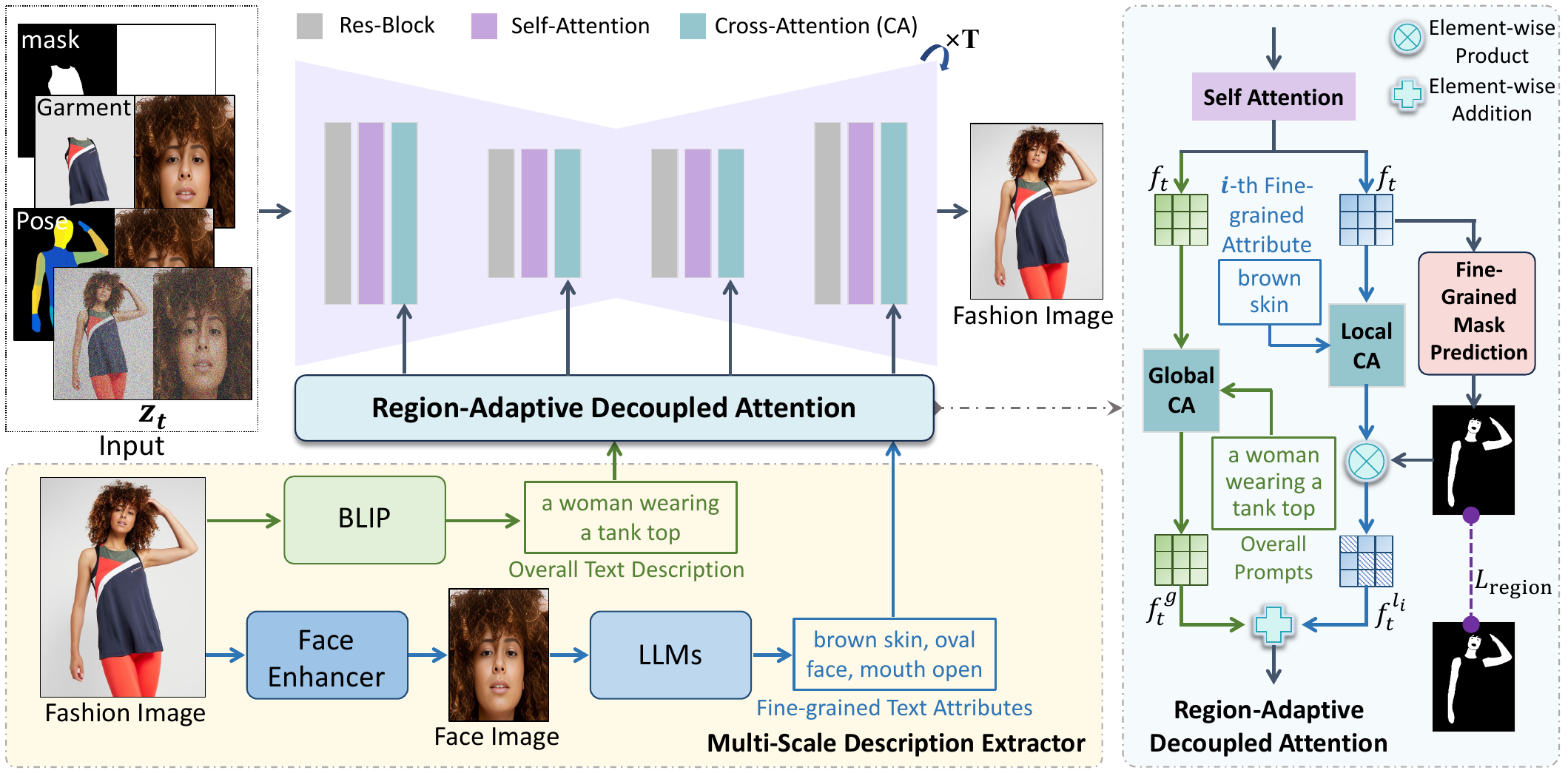}
   \caption{The overview of our framework. It consists of a multi-scale description extractor and a region-adaptive decoupled attention mechanism. 
   After training, our method can take facial images, overall showcase descriptions, or fine-grained text attributes as optional inputs to generate diverse showcase images.}
   \label{fig:pipeline}
\end{figure*}

\subsection{Human Image Generation}
Human image generation(HIG) task aims at synthesizing realistic and diverse human images.
We roughly categorize HIG into transfer-based and synthesis-based methods. Given source human images and target pose conditions, transfer-based algorithms~\cite{ma2017pose,han2023controllable,shen2023advancing,lu2024coarse} are expected to output photorealistic images with source appearance and target poses. In contrast, synthesis-based HIG methods~\cite{fruhstuck2022insetgan,fu2022stylegan,yang20233dhumangan,ju2023humansd,liu2023hyperhuman,wang2024towards,zhu2024mole,huang2024parts} concentrate on synthesizing high-quality human images conditioned on poses, text prompts, or faces. 
Early works on synthesis-based HIG are mainly based on GANs~\cite{fruhstuck2022insetgan,fu2022stylegan,yang20233dhumangan}, while many recent approaches have delved into designing specialized frameworks based on LDMs for better quality and controllability, such as incorporating more annotations~\cite{liu2023hyperhuman}, designing losses based on human-centric priors~\cite{ju2023humansd,wang2024towards}, or utilizing multiple human-part images~\cite{zhu2024mole}.  
Aside from these methods, a lot of personalized text-to-image methods~\cite{ye2023ip,li2024photomaker,shi2024instantbooth,peng2024portraitbooth} have been proposed to customize portraits conditioned on facial images.

\subsection{Garment-Centric Fashion Image Generation}
Garment-centric fashion image generation is a task to synthesize realistic fashion showcase images with specified garments under the guidance of images or text prompts.
MagicClothing~\cite{chen2024magic} is the first work focused on this area. It injects garment features into diffusion models to preserve details through self-attention feature integration. Parts2Whole~\cite{huang2024parts} proposes a reference-based framework that separately encodes multiple human appearance aspects (e.g., face, clothes, hair) from distinct images and employs multi-image conditioning and shared attention to compose a full-body output. IMAGDressing~\cite{shen2025imagdressing} generates fashion images using a hybrid architecture that combines a garment-specific UNet with a frozen diffusion backbone. DreamFit~\cite{lin2025dreamfit} features a lightweight encoder, which significantly reduces trainable parameters.
However, these methods typically require garment deformation in the generation, which often leads to garment texture distortions. Besides, they struggle to precisely control the fine-grained attributes of the generated models, due to the lack of specifically designed mechanisms.

\section{Approach}
In this section, we present FashionMAC, a deformation-free garment-centric fashion image generation framework for fine-grained model appearance customization. 
The proposed framework consists of two stages. In the first stage, given a garment image segmented from a dressed mannequin or a person, a garment-centric pose predictor is introduced to generate the corresponding pose that fits it. Please refer to the \textit{Supp.} for details. In the second stage, we build a deformation-free fashion generation model based on latent diffusion models (LDMs)~\cite{LDM}.

\subsection{Deformation-Free Fashion Generation Model}\label{sec:face}
In the second stage, we propose a deformation-free fashion generation model based on LDM to generate fashion showcase images conditioned on the given garment and the predicted pose maps, along with the text prompts and the facial images. The overall architecture is shown in \figref{fig:pipeline}. 

In this stage, the denoising U-Net has two kinds of input conditions: 1) conditions that are spatially aligned with the target fashion image, including the garment image, the pose map, and the garment mask indicating the regions to be outpainted, and 2) conditions that are not spatially aligned with the target fashion image, \ie the user-specified facial images of the synthesized fashion models. We leverage a simple yet effective feature fusion operation to deal with multiple conditions. For the conditions that are spatially aligned with the fashion image, we fuse them through channel-wise concatenation. 
For facial images that are not spatially aligned, we follow the self-attention-based texture transfer operation in~\cite{yang2024texture} and concatenate them with the fashion image in the spatial dimension, relying on the self-attention modules to decide which features to preserve. 

To enable precise fine-grained control over the generated model's appearance, we leverage information from both global descriptions of the complete fashion image and local attributes (skin tone, hairstyle, expression) that describe the appearance of the model, and explicitly introduce region-aware local structure guidance to ensure the controllability. To this end, we carefully design a multi-scale description extractor to parse a fashion image into hierarchical text prompts, region-adaptive decoupled attention mechanisms that perform global-local feature fusion under region-based spatial priors, and a chained mask injection strategy to propagate structural guidance across timesteps and provide more precise information. 

\subsection{Multi-Scale Description Extractor}
To provide effective textual guidance for controllable generation, the multi-scale description extractor (MDE) jointly captures global and fine-grained textual cues from a given fashion image. As shown in~\figref{fig:pipeline}, the extractor decouples the textual description into two levels — an overall prompt for the whole image and a set of fine-grained attributes focusing on local facial details.
To extract the overall global text prompt, we utilize a pretrained BLIP~\cite{li2022blip} model to caption the fashion image. For the fine-grained text prompts, we leverage a face enhancer module to obtain a high-quality facial image. The face enhancer first adopts Yolov5~\cite{ultralytics2021yolov5} to detect the face region of the fashion image. Then a face restoration model CodeFormer~\cite{zhou2022codeformer} and a super-resolution model RealESRGAN~\cite{wang2021real} are used in a sequential manner to obtain a high-quality facial image $F$. 
Since existing fashion datasets are short of detailed facial descriptions of the fashion models, the enhanced face image is fed into a large language model (LLM)~\cite{qwen2025qwen25technicalreport} in a captioning style, prompting it to output detailed appearance attributes such as skin tone, facial shape, expression,~\etc. These form the fine-grained attribute set that is used for localized guidance during generation.

\subsection{Region-Adaptive Decoupled Attention}
In diffusion-based text-to-image models, the cross-attention layers are responsible for linking textual attributes with corresponding visual regions. 
However, in practice, certain regions of the denoising feature map may fail to effectively respond to their associated textual tokens. 

To further improve the cross-attention layers to better align the visual concepts with fine-grained text attributes, we propose Region-Adaptive Decoupled Attention (RADA) — a dual-branch attention design that explicitly decouples global and fine-grained attribute information and selectively adapts attention responses based on region-aware priors. It is utilized to replace each cross-attention layer of the denoising U-Net (including the encoder and the decoder), as illustrated in~\figref{fig:pipeline}. Given an overall global prompt $\tau_g$, a fine-grained prompt $\tau_l$ and the denoising feature $f_t$ output by a self-attention layer at timestep $t$, RADA instantiates two separate cross-attention branches: the Global Cross Attention GCA attends to the overall prompt embedding and the Local Cross Attention LCA attends to individual fine-grained attribute embeddings. 

In each local branch, we utilize a lightweight mask prediction head to provide spatial structural priors to guide the attention mechanism during generation. The core idea is to predict soft region masks for each local attribute, indicating where in the image the model should focus its attention for better text-image alignment. Each head has three stacked $3\times3$ convolution layers followed by a sigmoid activation. We take the output features from the self-attention layer and feed them into the prediction head to generate a set of masks with $N$ channels and resolution matching that block’s feature size, where $N$ is the total number of fine-grained attribute classes. We apply a region-level supervision loss $L_{region}$ to optimize the heads for accurate masks.
These masks act as explicit spatial cues to enforce structural information in the global-local fusion process. 

Then RADA can be represented as:
\begin{equation}
\resizebox{0.9\linewidth}{!}{%
    $RADA(f_t, \tau_g,\tau_l ) = GCA(f_t, \tau_g) + \sum_{i}^{N} LCA(f_t,\tau_{l_i}) \odot M^i_{t/t-1} .$
}
\end{equation}
where $i$ is the $i$-th fine-grained attribute, $M^i_{t/t-1}$ is the corresponding mask generated by the prediction head at timestep $t$ or $t-1$, which will be exlain in the following section. $\odot$ represents the element-wise product operator. In this way, RADA introduces explicit structural priors into the diffusion attention mechanism by decoupling global and local semantics and leveraging attribute-specific spatial masks. This improves the visual-textual alignment and enhances the fine-grained features. 

\subsection{Chained Mask Injection Strategy}
To enhance the structural controllability of image generation, we explore the injection of region-level guidance at different stages of the denoising process. 
Our empirical findings suggest that incorporating region structure priors across all timesteps consistently improves generation quality. Interestingly, we observe that the benefits are particularly pronounced during early timesteps, where the model is more sensitive to structural cues. This observation aligns with insights from prior work such as Prospect~\cite{zhang2023prospect}, which points out that in the early timesteps of the denoising process, diffusion models primarily focus on recovering global structure and semantic layout.

Furthermore, we investigate the injection locations within the network architecture and find that applying the region priors to both encoder and decoder blocks across all layers yields the best overall performance. However, this introduces a practical challenge: the encoder blocks, being relatively shallow, struggle to generate reliable region masks from highly noisy features in the early denoising steps. In contrast, decoder blocks benefit from deeper feature integration and are capable of producing reasonably accurate masks even under noise conditions.

\begin{figure}[t]
  \centering
  \includegraphics[width=\linewidth]{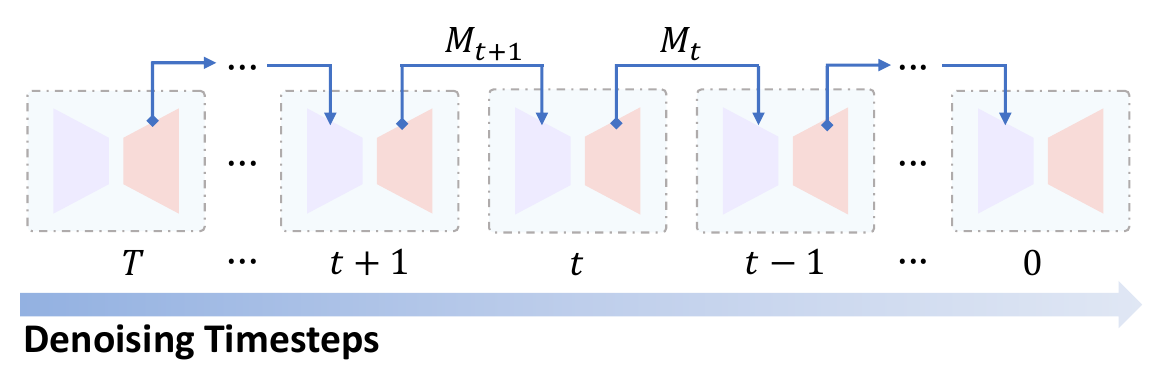}
   \caption{The chained mask injection strategy. }
   \label{fig:cmi}
\end{figure}

To fully leverage the potential of the region priors, we propose a chained mask injection 
(CMI) strategy. During inference, we utilize the predicted mask $M_t$ from the last decoder block at timestep $t$ to guide all encoder blocks at timestep $t-1$ instead of predicting masks from scratch, as illustrated in~\figref{fig:cmi}. In the meantime, for decoder blocks, the region-aware RADA module uses the mask predicted from its corresponding decoder block, since predicted masks from the decoder tend to be refined during the denoising. This chained temporal connection effectively bootstraps the encoder with decoder-refined structure priors, allowing both parts of the network to benefit from more accurate region guidance.

\subsection{Training Objective}
The training objective of the deformation-free fashion generation model comprises two components: a denoising loss $\mathcal{L}_{denoise}$ for the U-Net $\epsilon_\theta$  and a region loss $\mathcal{L}_{region}$ for supervising the fine-grained mask prediction in RADA. 

For the denoising loss, we follow the standard training paradigm of LDM, where the network is trained to predict the added noise given a noisy latent $\mathbf{z_t}$ and conditional inputs $\mathbf{c}$. Specifically, we minimize the mean squared error between the predicted noise and the ground truth noise:
\begin{equation}
\mathcal{L}_{denoise} = \mathbb{E}_{\mathbf{z},\epsilon\sim\mathcal{N}(0,1),t} [ || \epsilon - \epsilon_{\theta} (\mathbf{z_t},t,\mathbf{c}) ||_2^2].
\end{equation}

For the region supervision, we simply leverage a Euclidean loss to train spatial masks. Given  ground truth attribute region masks $\{M^i\}_{i=1}^N$, and predicted masks $\{\hat{M}^i_t\}_{i=1}^N$ of timestep $t$ at resolution corresponding to each decoder block, the loss is employed over all $N$ channels:

\begin{equation}
    \mathcal{L}_{region }=\frac{1}{N} \sum_{i=1}^{N}\left\| {\hat{M}^i_t}-M^i\right\|_{2}^{2} .
\end{equation}

\begin{figure*}[t]
  \centering
  \includegraphics[width=0.94\linewidth]{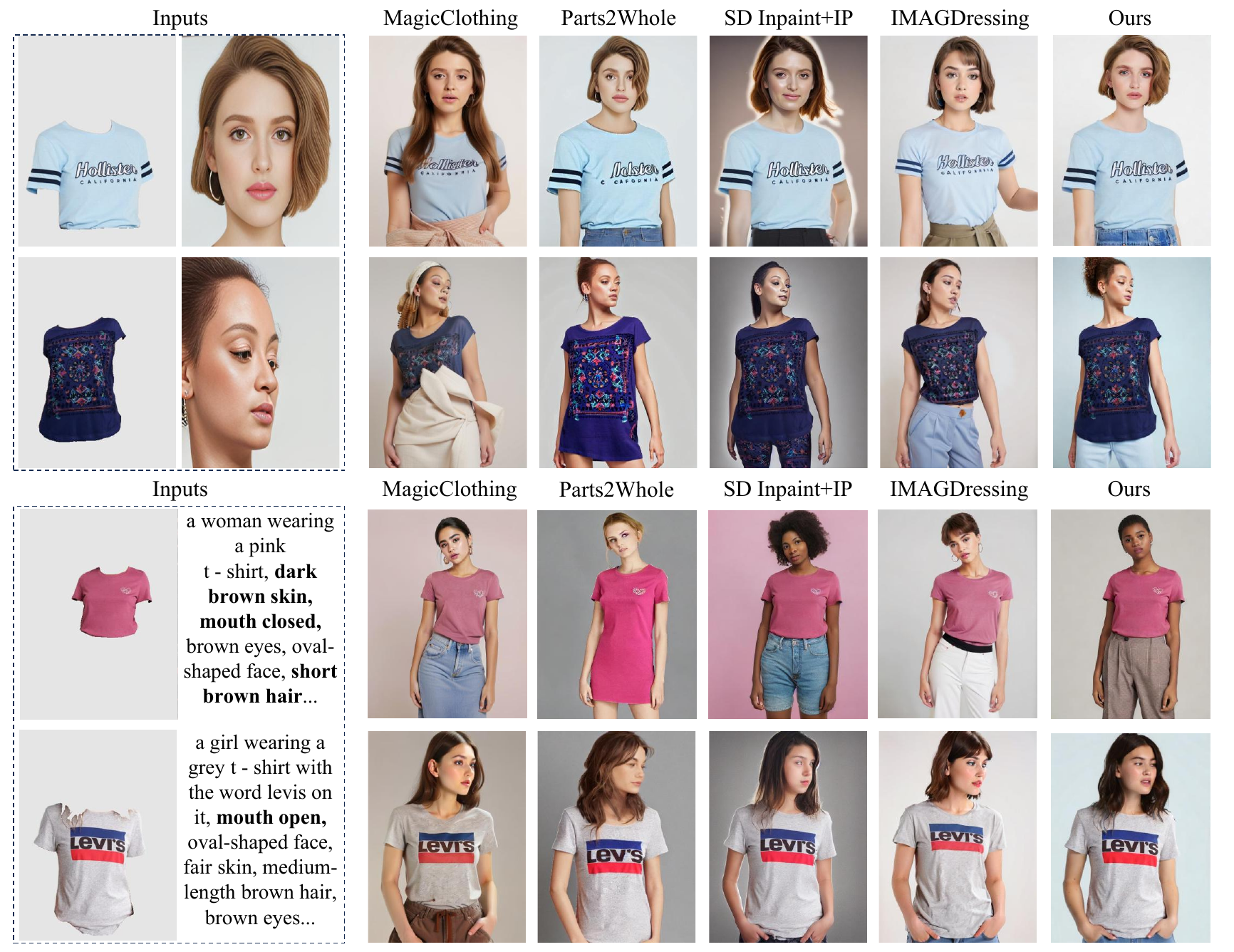}
   \caption{Comparison with the baseline methods. The first two rows show the results with facial image guidance. The last two rows show results with text prompt guidance. }
   \label{fig:results}
\end{figure*}

\section{Experiments}
\subsection{Implementation Details}
We conduct experiments on the virtual try-on benchmark VITON-HD~\cite{choi2021viton}, which contains $14,221$ training images and $2,032$ testing images. The model is trained at $576\times768$. FashionMAC is built on Stable Diffusion 1.4~\cite{LDM}. All experiments are implemented with Pytorch and performed on $4$ NVIDIA A100 80G GPUs. Please refer to the \textit{Supp.} for more information and additional results on the IGPair dataset\cite{shen2025imagdressing}.

\begin{table*}
\small
\centering
\begin{tabular}{ccccccccc} 
\toprule
\multirow{2}{*}{Methods} & \multicolumn{4}{c}{With facial image guidance}                                                                        & \multicolumn{4}{c}{With text prompts guidance}                                                       \\ 
\cmidrule(l){2-9}
                         & FID$\downarrow$ & KID$\downarrow$  & CLIP-i$\uparrow$       & MP-LPIPS$\downarrow$  & FID$\downarrow$ & KID$\downarrow$ & CLIP-t$\uparrow$ & MP-LPIPS$\downarrow$  \\ 
 \midrule
SD Inpaint               & 58.58           & 0.016          & 0.83                   & 0.119                                             & 40.32           & 0.0186          & 0.224            & 0.080                                \\
Parts2Whole              & 41.76           & 0.031          & 0.89                   & 0.128                                             & 41.17           & 0.0284          & 0.214            & 0.120                                \\
MagicClothing            & 41.74           & 0.030          & 0.85                   & 0.117                                              & 37.24           & 0.0303          & 0.228            & 0.117                                \\
IMAGDressing             & 29.34           & 0.021          & 0.89                   & 0.083                                            & 27.80           & 0.0204          & 0.227            & 0.080                                \\
Ours                     & \textbf{14.89}  & \textbf{0.008} & \textbf{0.95} & \textbf{0.055}                   & \textbf{13.68}  & \textbf{0.0060} & \textbf{0.231}   & \textbf{0.051}                       \\
\bottomrule
\end{tabular}
\caption{Quantitative comparison with different methods.}
\label{tab:comparison}
\end{table*}

\subsection{Comparisons}
\paragraph{Baselines.} 
We compare FashionMAC with the related inpainting-based and garment-driven image synthesis approaches, including SD Inpaint~\cite{LDM}, Parts2Whole~\cite{huang2024parts}, MagicClothing~\cite{chen2024magic}, and IMAGDressing~\cite{shen2025imagdressing}.

\paragraph{Metrics.} To evaluate the methods, we calculate the metrics in two settings: with facial images and with text prompts only. We employ LPIPS~\cite{zhang2018unreasonable}, FID~\cite{heusel2017gans}, KID~\cite{bińkowski2018demystifying} and DreamSim~\cite{fu2023dreamsim} to measure the realism of the generated images from different dimensions.
We adopt CLIP-i and CLIP-t~\cite{radford2021learning} to estimate the similarity of the CLIP space between the generated images and the ground truths, MP-LPIPS~\cite{chen2024magic} to evaluate whether the characteristics of the target garment are well-preserved.  

\paragraph{Qualitative Results.}
\figref{fig:results} demonstrates the qualitative comparisons between FashionMAC and state-of-the-art baseline methods. The first two rows show the results with facial image guidance, while the last two rows present the results with only text prompts. 
We observe that our method achieves better results with fewer artifacts in both settings. 
On one hand, as FashionMAC utilizes a deformation-free outpainting-based framework, the garment details, especially the characters and patterns, are better preserved. In contrast, the garment deformation in baseline methods may introduce undesired distortions or color deviation. FashionMAC can also maintain facial identity and structural integrity more accurately with the face guidance.
On the other hand, our method outperforms others in terms of both visual quality and appearance controllability for the comparisons with only text prompts. The baseline methods tend to ignore some of the attributes, such as `dark brown skin',  `mouth open', and `dark brown hair'. In contrast, FashionMAC is capable of maintaining the semantic alignments between the generated fashion images and the specified prompts. 

\paragraph{Quantitative Results.}
\tabref{tab:comparison} presents the quantitative comparison results between FashionMAC and the baseline methods. Our method clearly outperforms existing state-of-the-art methods with large margins on all the metrics under the guidance of either the facial images or text prompts. Among these metrics, MP-LPIPS numerically validates the effectiveness of FashionMAC for faithful preservation of the intricate garment details. The results on FID and KID verify the superiority of our method in terms of image quality and realism. 
The CLIP-i score of our method indicates that FashionMAC accurately synthesizes fashion images, while the CLIP-t score demonstrates that our method achieves better visual-textual alignment than baseline methods.

\begin{table}[t]
\centering
\resizebox{0.9\linewidth}{!}{
\begin{tabular}{cccc} 
\toprule
Method            &  FID↓                   & KID↓                  & CLIP-t↑                 \\ 
\midrule
FashionMAC   & \textbf{13.68}  &\textbf{0.0060}  & \textbf{0.231} \\ 
w/o CMI    & 14.91  & 0.0073 & 0.221  \\  
w/o CMI and RADA  & 15.21  & 0.0077 & 0.222  \\
\bottomrule
\end{tabular}
}
\caption{The quantitative results of the ablation study. }
\label{tab:ablation}
\end{table}

\subsection{Ablation Study}
\begin{figure}[t]
    \centering
    \includegraphics[width=\linewidth]{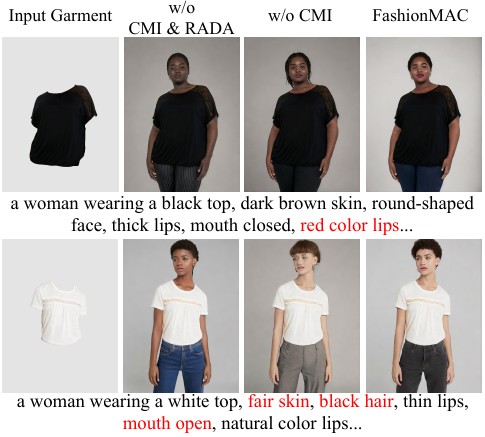}
    \caption{The qualitative results of the ablation study.}
    \label{fig:ablation}
\end{figure}

To assess the effectiveness of the proposed Chained Mask Injection (CMI) strategy and Region-Adaptive Decoupled Attention (RADA) module, we conduct ablation studies on our full model FashionMAC, model without CMI and model without CMI and RADA. As shown in~\tabref{tab:ablation}, removing the CMI strategy results in performance degradation across all metrics including FID, KID and CLIP-t, indicating that temporal mask propagation significantly enhances generation quality and text-image alignment. Further removing the RADA module still leads to drops in visual fidelity (FID to 15.21, KID to 0.0077), demonstrating that incorporating spatially aligned region priors is critical for guiding the denoising process. The ~\figref{fig:ablation} demonstrates the qualitative results. Removing the CMI strategy degrades the structural accuracy and visual fidelity. Further removing the RADA module leads to more misalignments between the generated image and the input prompts (e.g. skin color and mouth status). These results validate the complementary roles of CMI and RADA in improving both visual realism and controllability. 

\section{Conclusion}
In this work, we presented FashionMAC, a novel garment-centric fashion image generation framework that eliminates the need for garment deformation and enables fine-grained appearance customization. By directly generating fashion showcase images from dressed garment inputs, our method leverages readily available visual data while avoiding distortion issues introduced by garment deformation. 
To achieve fine-grained appearance controllability over the synthesized human models (e.g., hairstyle, skin tone, expression,~\etc ), we propose a Region-Adaptive Decoupled Attention (RADA) mechanism coupled with a Chained Mask Injection (CMI) strategy. RADA selectively modulates cross-attention responses based on region-aware priors predicted by fine-grained mask prediction, while CMI progressively propagates structural guidance across timesteps to enhance spatial precision.
Extensive experiments demonstrate that FashionMAC outperforms existing baselines in both visual fidelity and appearance controllability, offering a practical solution for e-commerce scenarios.

\section{Acknowledgements}
This work was partially supported by Zhejiang Provincial Natural Science Foundation of China (No. LQ23F020009, No. LQN25F020012, No. LD24F020011) and NSFC (No. 62302449, No. 62402439, No. 92570206, No. 62421003).

\bibliography{aaai2026}

\clearpage

\oldtwocolumn[
\begin{center}  
    \LARGE\bfseries
    Supplementary Material for FashionMAC: Deformation-Free Fashion Image Generation
with Fine-Grained Model Appearance Customization
    \vspace{1.25em}
\end{center}
]



In this supplementary material, we provide more details of the FashionMAC and additional results.

\section{Garment-Centric Pose Predictor}\label{sec:pose}
Without any human structural priors, directly outpainting a deformed garment to synthesize a fashion model wearing it is very difficult. Therefore, we first design a garment-centric pose prediction model to generate the corresponding poses that fit the given garment.  

Specifically, the garment-centric pose predictor is based on the LDM framework. Given an input fashion showcase image, we obtain the deformed garment image and the corresponding pose map from it by a pre-trained cloth segmentation model~\cite{dabhi2021clothsegmentation} and a DensePose model~\cite{guler2018densepose}, respectively. We then train a UNet denoiser $\epsilon_{\theta_p}$ from scratch to generate the pose conditioned on the garment. Since the pose map and the garment image are spatially aligned, we simply concatenate the garment image's latent encoding $z_c$ with the pose image's noised latent encoding  ${z_p}_t$ in the channel dimension as the input to the UNet denoiser $\epsilon_{\theta_p}$ at timestep $t$. The objective function for training is as follows:
\begin{equation}
\mathcal{L}_{pose} = \mathbb{E}_{p, c,\epsilon\sim\mathcal{N}(0,1),t} [ || \epsilon - \epsilon_{\theta_p} ({z_p}_t, z_c, t) ||_2^2].
\end{equation}

During the inference, thanks to the stochasticity of the reverse denoising process, we can utilize the garment-centric pose predictor to sample diverse pose maps that fit the given garment.

\section{Dataset}
To further evaluate the generalization ability of our method, we conduct experiments on the IGPair~\cite{shen2025imagdressing} dataset. The original dataset contains over $300,000$ training pairs and $2,000$ testing pairs. To adapt it to our task, we construct a curated subset. Specifically, we select six garment categories based on the labels, including `upper\_body', `dresses', `skirts', etc. We then perform de-duplication to ensure that each garment image corresponds to a single fashion showcase image. 
To guarantee the quality of facial generation, we filter the dataset to retain images with a sufficient face resolution (specifically, a face area ratio between $0.011$ and $0.05$). Furthermore, we adjust the pose distribution by decreasing the proportion of images with side or back views. Consequently, we obtain a dataset comprising $24,628$ training pairs and $2,000$ testing pairs for our experiments. The models are trained at a resolution of $512\times640$.

To support our framework, we construct fine-grained text prompts and corresponding spatial masks for each training image of our two datasets. 
For the prompt extraction, we adopt QWen 2.5VL~\cite{qwen2025qwen25technicalreport} to generate descriptive attribute prompts for predefined semantic regions. We define 9 core region categories that are most relevant to fashion, including expression, skin, hair, mouth, etc.
To extract the spatial region masks, we utilize the semantic segmentation maps from our datasets for the body region and employ additional segmentation tools SAM~\cite{kirillov2023segment} and Bisenet~\cite{yu2018bisenet} to obtain the hair and face region.

\section{Implementation Details}

We adopt a three-stage training strategy: 1) We train FashionMAC without RADA for $55,000$ iterations, using a batch size of $16$ and learning rate of $2e-5$. 2) We add the RADA module and continue finetuning with ground-truth region masks for $6,000$ iterations, using the same learning rate. 3) We freeze the denoising model and RADA, and train the mask prediction head for $60$ epochs with a learning rate of $1e-4$.
In inference, we replace ground-truth masks with predicted ones and apply the chained mask injection strategy to propagate structural guidance across timesteps.

\section{Results Conditioned on Virtual Faces}
\begin{figure}[h]
    \centering
    \includegraphics[width=1\linewidth]{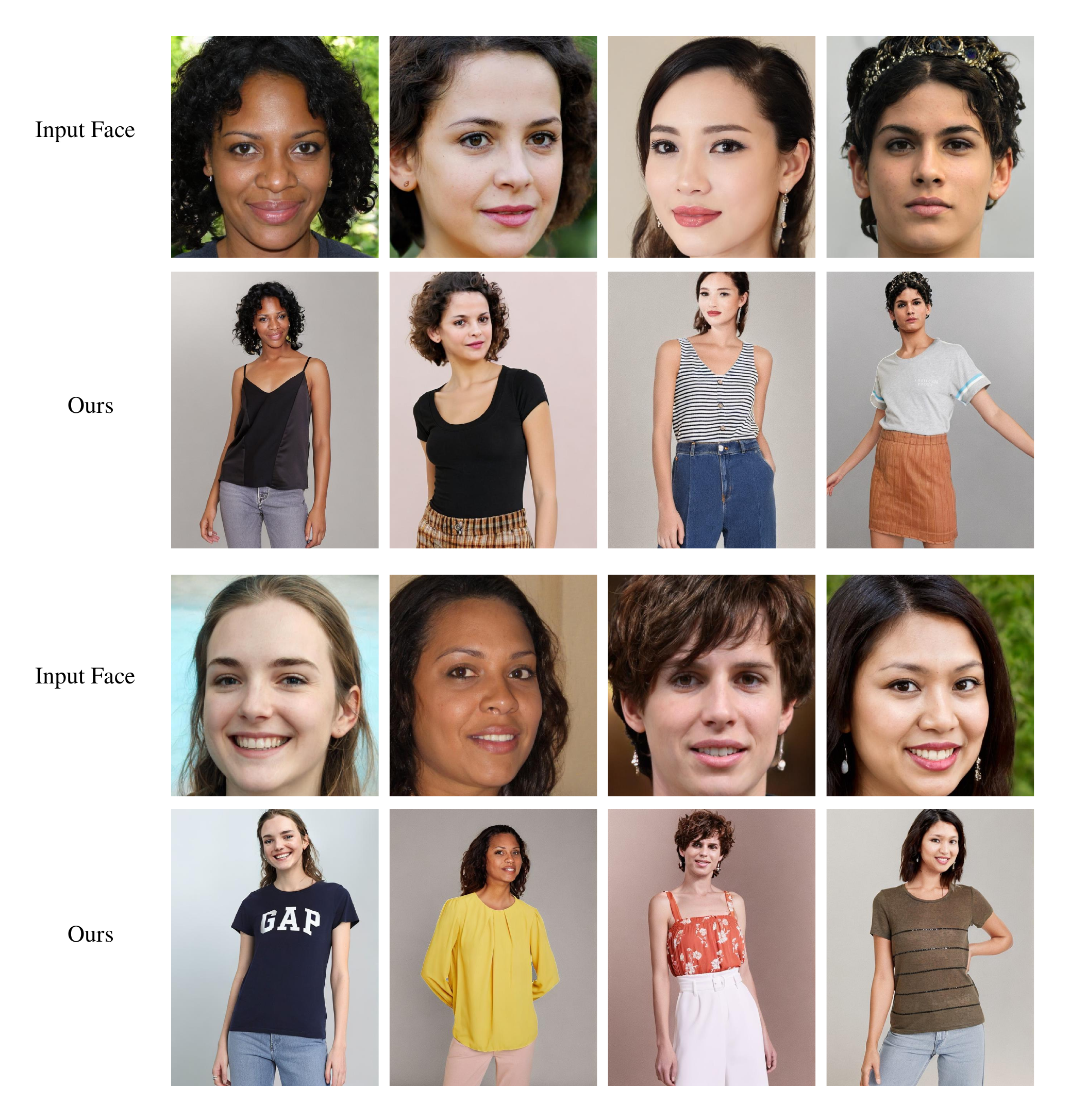}
    \caption{The results of our method conditioned on virtual facial images generated by StyleGAN.}
    \label{fig:stylegan}
\end{figure}

Our FashionMAC can generate fashion showcase images under the guidance of facial images. To avoid issues of portraiture rights of real humans, the input facial images can be obtained from portrait generation approaches such as StyleGAN~\cite{karras2019style}. ~\figref{fig:stylegan} demonstrates the results of FashionMAC conditioned on virtual faces generated by StyleGAN. The generated images successfully preserve the distinctive characteristics of the input faces, showcasing FashionMAC's ability to maintain high fidelity in face-conditioned image generation.

\section{Qualitative Results of the Ablation Study}

\begin{figure*}[!h]
    \centering
    \includegraphics[width=\linewidth]{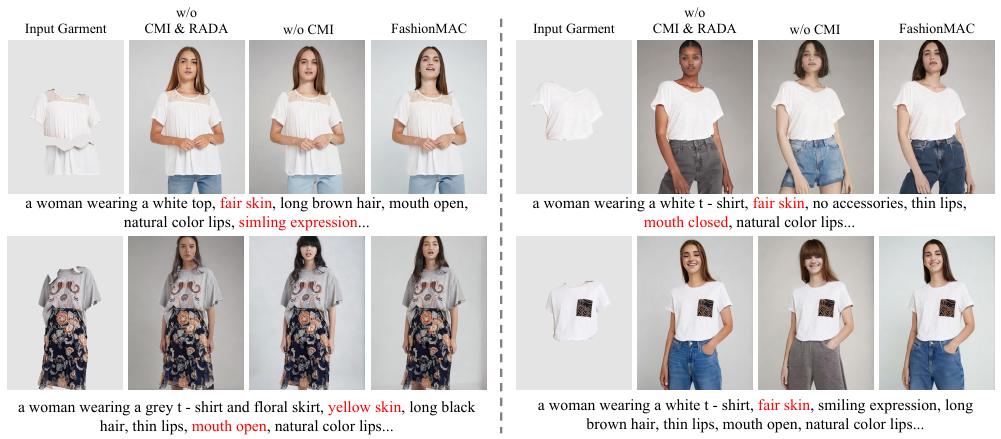}
    \caption{The qualitative results of the ablation study.}
    \label{fig:x_ablation}
\end{figure*}

More qualitative results of the ablation study are shown in ~\figref{fig:x_ablation} to assess the effectiveness of the proposed Chained Mask Injection (CMI) strategy and Region-Adaptive Decoupled Attention (RADA) module. The ablation studies are conducted on our full model FashionMAC, model without CMI and model without CMI and RADA. These results further validate the effectiveness of CMI and RADA.

\section{Experiments of the Mask Prediction }

\begin{figure*}[h]
    \centering
    \includegraphics[width=0.9\linewidth]{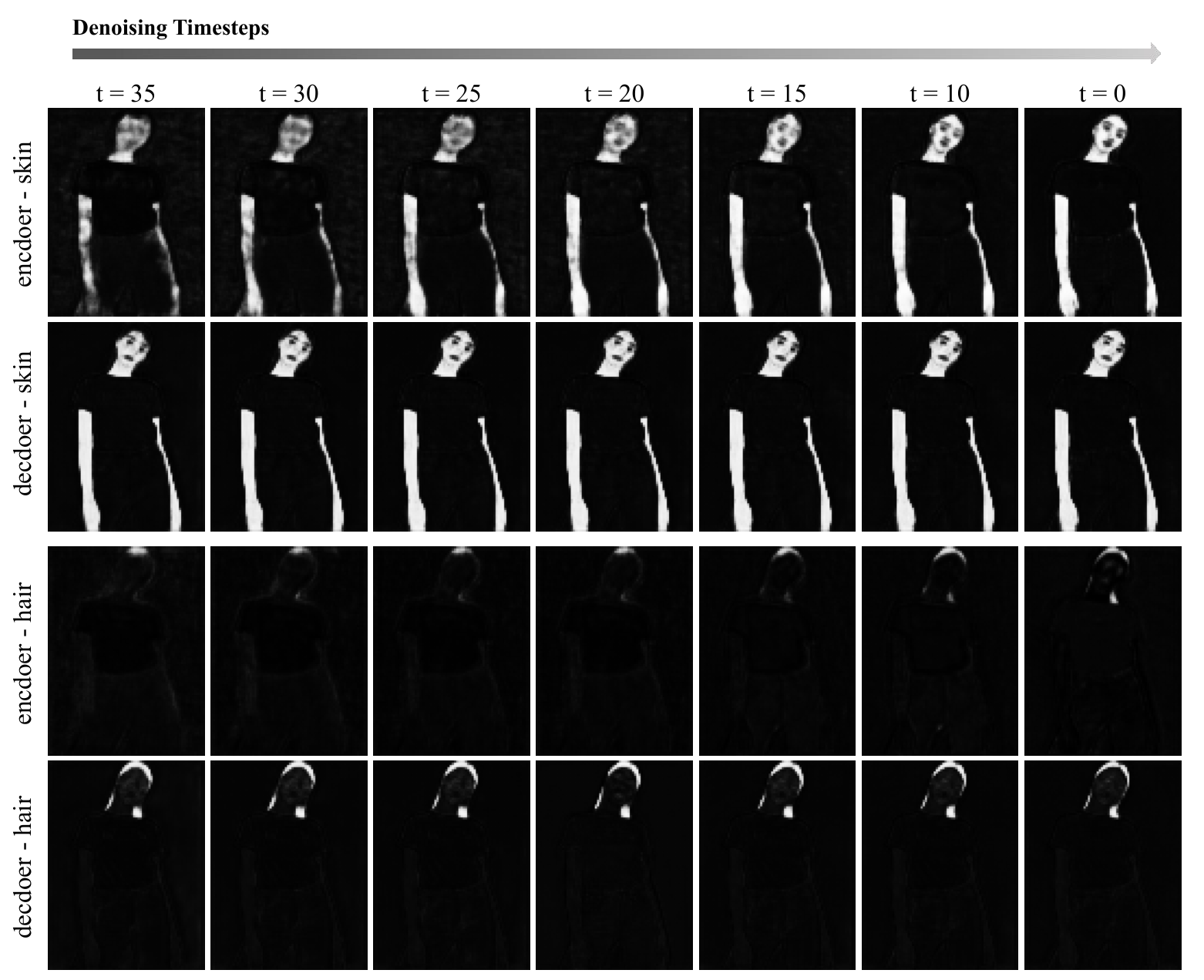}
    \caption{The masks predicted by different parts of the network at different timesteps.}
    \label{fig:mask-prediction}
\end{figure*}

To enable accurate region-aware conditioning during the denoising process, we explore where to place the mask prediction head for optimal spatial guidance. We experimentally found that applying mask prediction on low-level encoder features at early denoising timesteps tends to produce imprecise region masks. In contrast, decoder-based masks remain relatively stable and less affected by the timestep. The results are illustrated in~\figref{fig:mask-prediction}. 
We attribute this to the noisy nature of early-step features and the limited capacity of shallow encoder blocks to extract meaningful spatial cues from them. Conversely, decoder features benefit from deeper hierarchical feature integration, which enhances their ability to localize fine-grained semantic regions. Based on this observation, we adopt a chained mask injection strategy that progressively predicts and injects masks at network layers across timesteps, providing stable and precise regional priors to guide generation. The mask generation introduces only a marginal computational overhead—approximately $32.7$ MFlops, corresponding to a $7\%$ increase in inference time.

\section{Robustness Analysis of Segmentation}
The primary goal of FashionMAC is to generate high-quality garment showcase images for e-commerce retailers, rather than performing traditional Virtual Try-On (VTON). The critical objectives of this task are (i) preserving garment details and (ii) enabling controllable model appearance customization. Unlike standard VTON methods, our approach does not require generating garments from scratch; instead, it utilizes garment images directly segmented from the source models.Thanks to the powerful capabilities of modern segmentation models (e.g., SAM 3~\cite{carion2025sam3segmentconcepts}), our method receives reliable semantic conditioning in most cases. Furthermore, for occasional imperfect segmentations, users can manually improve the semantic maps using editing tools (e.g., Photoshop) to mitigate potential artifacts.

\begin{figure*}
    \centering
    \includegraphics[width=0.9\linewidth]{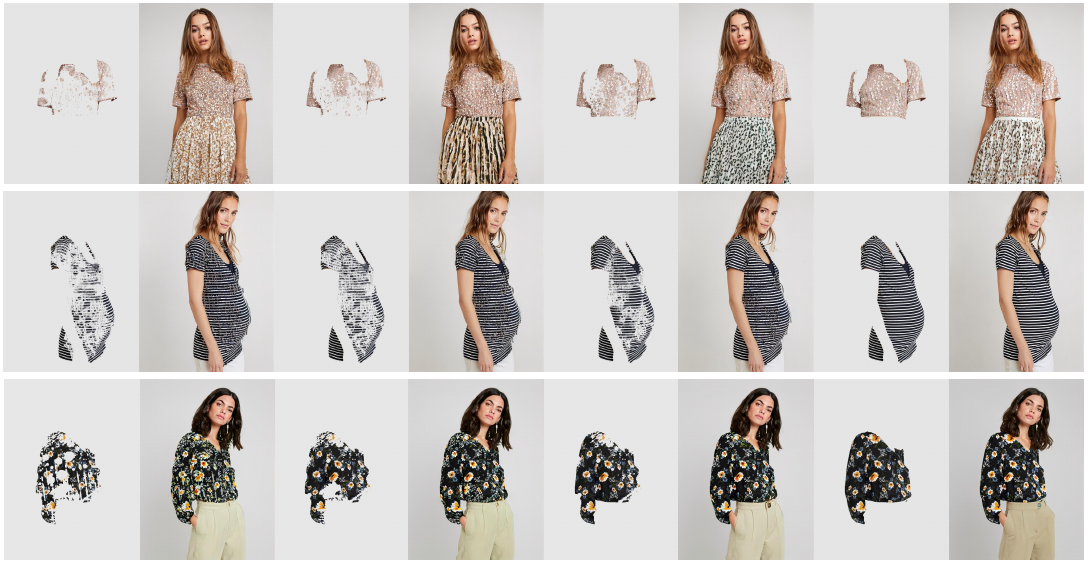}
    \caption{Results for the same garment under varying segmentation quality.}
    \label{fig:robustness_result}
\end{figure*}

To further evaluate the impact of segmentation quality on our results, we conducted experiments by degrading the input garment images to simulate varying degrees of segmentation artifacts. The results are illustrated in~\figref{fig:robustness_result}. We observe that even under extreme cases, our method is capable of maintaining the overall appearance to some extent. However, faithfully preserving specific details within the missing regions remains challenging. As the segmentation quality increases, the alignments between the generated images and the target garment are improved.




\section{More Results of the Fine-grained Customization}

\begin{figure*}
    \centering
    \includegraphics[width=0.85\linewidth]{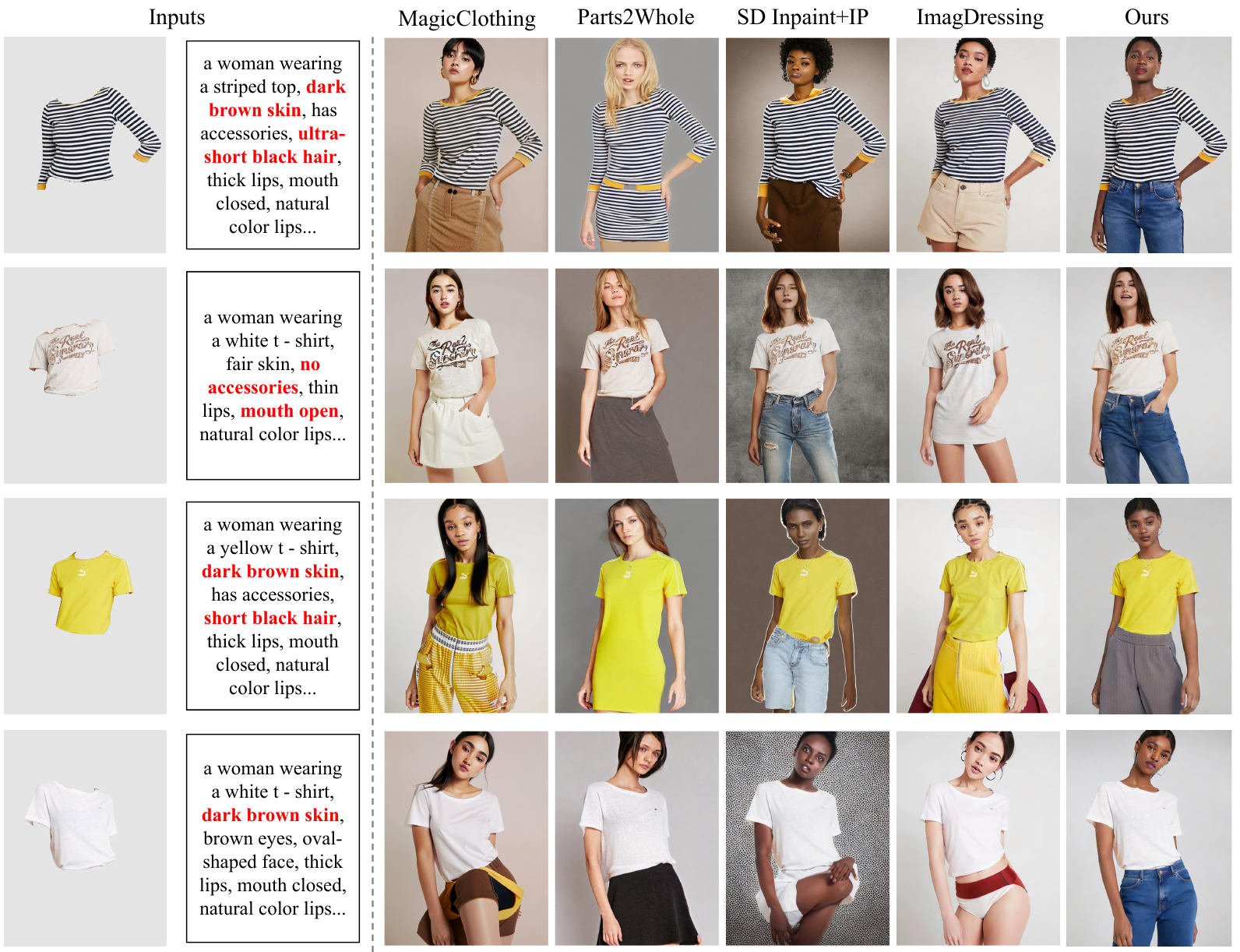}
    \caption{Qualitative comparison with existing methods.}
    \label{fig:more_comparison}
\end{figure*}

\begin{figure*}
    \centering
    \includegraphics[width=0.85\linewidth]{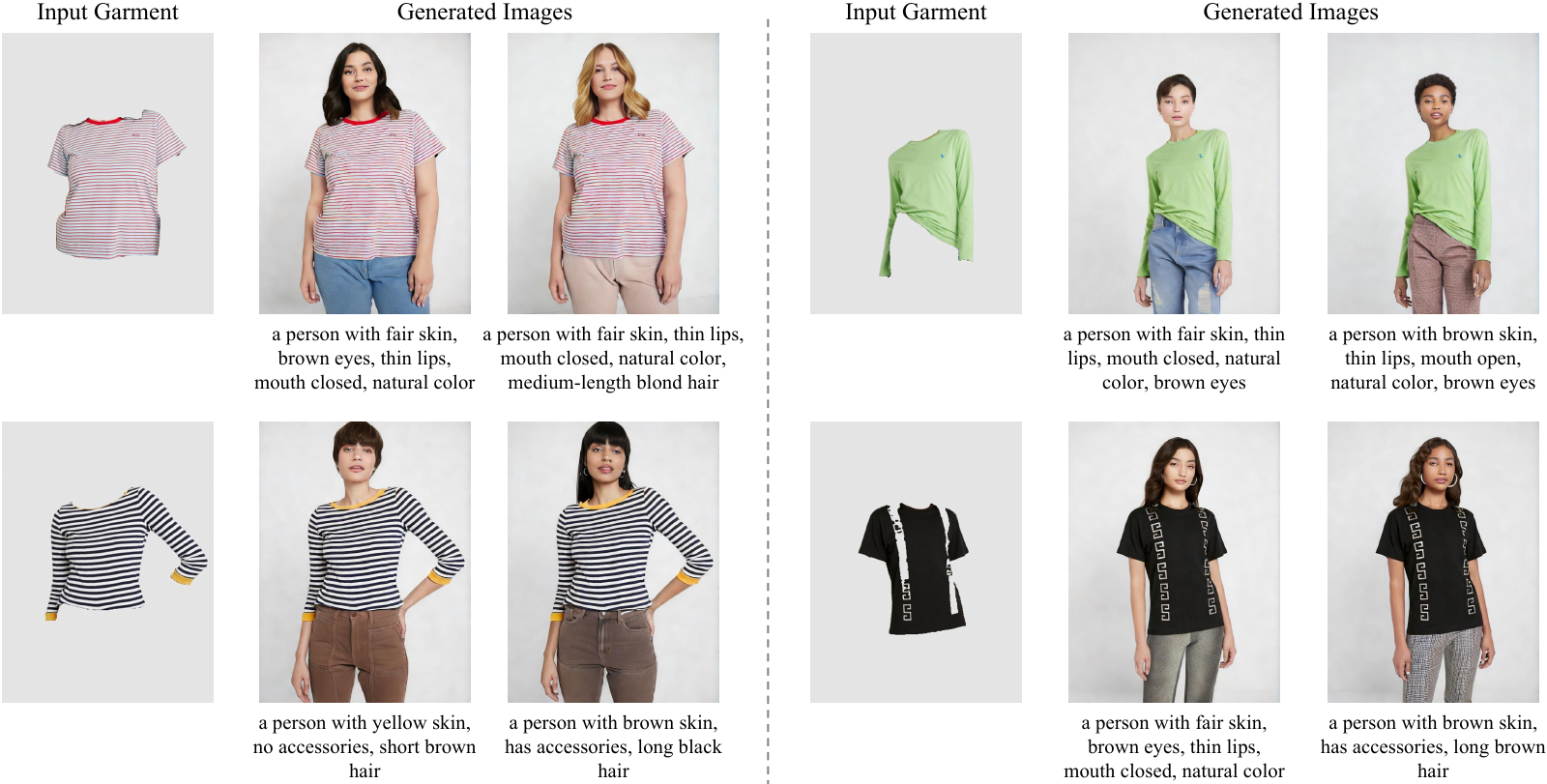}
    \caption{More Results of FashionMAC.}
    \label{fig:more_results}
\end{figure*}

\figref{fig:more_comparison} and \figref{fig:more_results} present additional qualitative comparisons and diverse visual results of FashionMAC. These examples highlight the framework’s superior capability for fine-grained appearance customization and its effectiveness in generating high-fidelity, diverse fashion showcase images.

\begin{table*}
\small
\centering
\begin{tabular}{c cccc ccc} 
\toprule
\multirow{2}{*}{Methods} & \multicolumn{4}{c}{With facial image guidance} & \multicolumn{3}{c}{With text prompts guidance} \\ 
\cmidrule(l){2-8}
& FID$\downarrow$ & KID$\downarrow$ & CLIP-i$\uparrow$ & MP-LPIPS$\downarrow$ & FID$\downarrow$ & KID$\downarrow$ & MP-LPIPS$\downarrow$ \\ 
\midrule
MagicClothing & 39.17 & 0.0213 & 0.81 & 0.178 & 23.36 & 0.0086 & 0.167 \\
IMAGDressing & 20.27 & 0.0040 & 0.89 & 0.164 & 16.97 & 0.0047 & 0.155 \\
Ours & \textbf{10.23} & \textbf{0.0026} & \textbf{0.96} & \textbf{0.146} & \textbf{13.10} & \textbf{0.0040} & \textbf{0.146} \\
\bottomrule
\end{tabular}
\caption{Quantitative comparison on the IGPair dataset.} 
\label{tab:igpair_metrics}
\end{table*}


\section{Generation Conditioned on Hard-set Masks.}
\begin{figure*}
    \centering
    \includegraphics[width=0.85\linewidth]{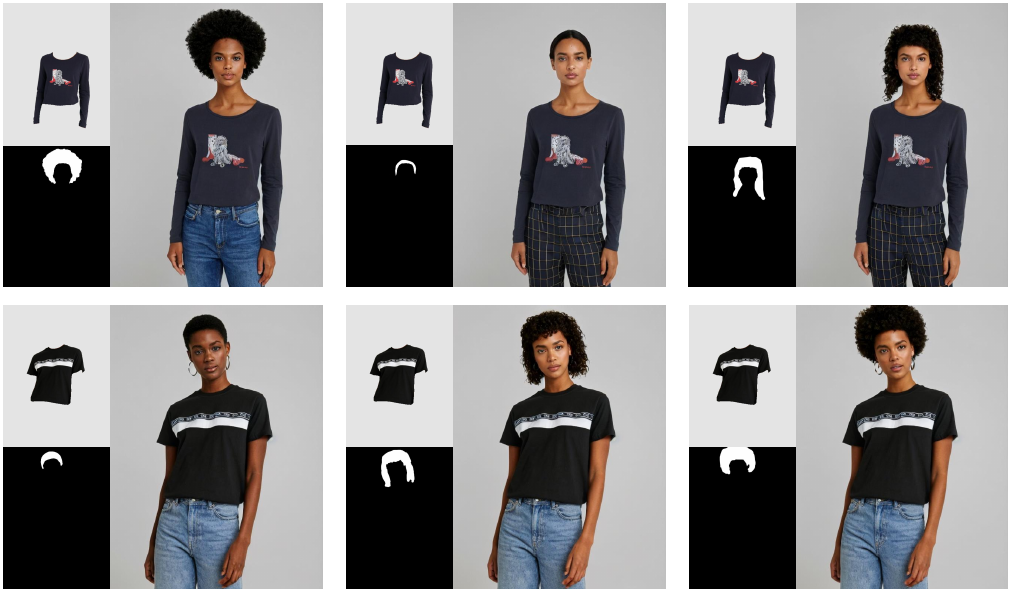}
    \caption{The results conditioned under hard-set masks for the hair region.}
    \label{fig:hard_mask}
\end{figure*}

Our model can generate fashion images conditioned on hard-set spatial masks. The results are presented in \figref{fig:hard_mask}, which demonstrates the ability of FashionMAC to faithfully adhere to the specified region during generation. It further validates the effectiveness of our proposed Region-Adaptive Decoupled Attention (RADA) strategy, confirming that FashionMAC is capable of supporting fine-grained, customized image generation.

\section{Experiments on the IGPair Dataset}


\paragraph{Baselines.}
We compare FashionMAC with MagicClothing~\cite{chen2024magic} and IMAGDressing~\cite{shen2025imagdressing} on the selected IGPair subset. To ensure a fair comparison, we retrain the baseline methods on the same IGPair subset.

\begin{figure*}
    \centering
    \includegraphics[width=1.0\linewidth]{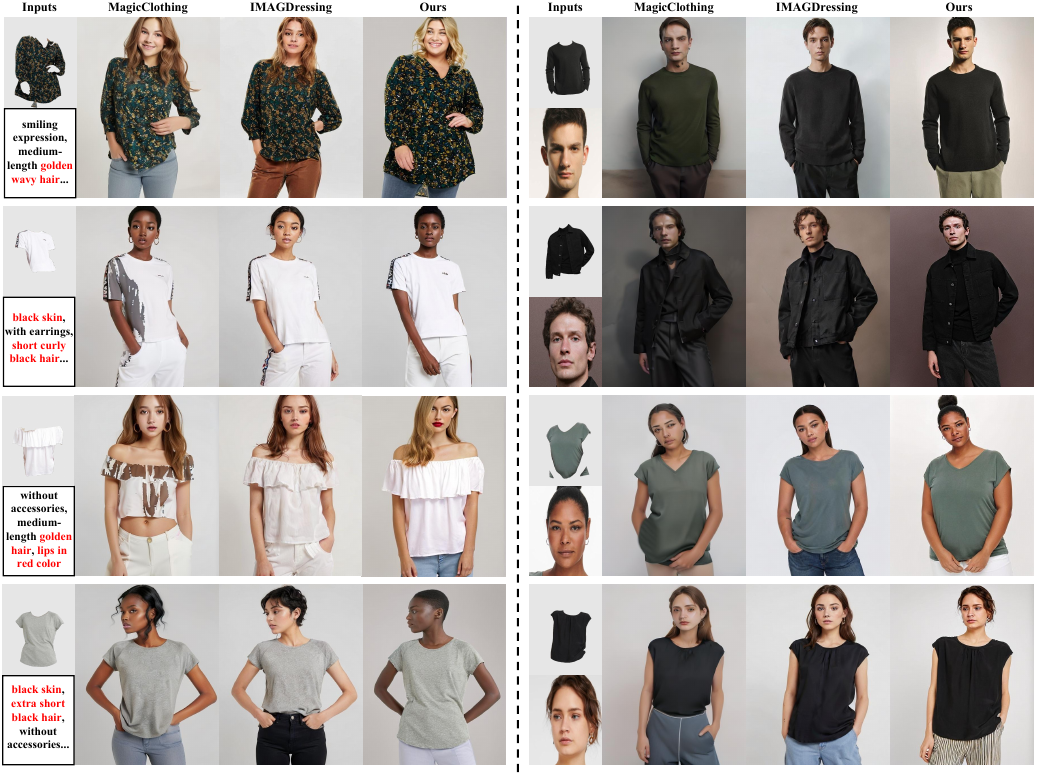}
    \caption{Comparison with the baseline methods on the IGPair subset. The left side shows the results with text prompt guidance. The right side shows the results with facial image guidance.}
    \label{fig:IGPair_compairsion}
\end{figure*}

\paragraph{Comparison.}
\figref{fig:IGPair_compairsion} presents the qualitative comparisons between FashionMAC and the baseline methods. The left panel displays the results generated with only text prompts, while the right panel shows the results under facial image guidance. For the results conditioned solely on text prompts, our method achieves superior text-image consistency compared to the baselines, while simultaneously maintaining high garment fidelity. Regarding the results under facial image guidance, our method demonstrates the ability to faithfully preserve the given facial identity.
\tabref{tab:igpair_metrics} shows the quantitative comparison results between FashionMAC and the baseline methods. Our method outperforms MagicClothing and IMAGDressing across all evaluation metrics. These results further demonstrate the superior capability of FashionMAC.


\section{Comparison with the Virtual Try-On Based Methods}

\begin{figure*}
    \centering
    \includegraphics[width=0.85\linewidth]{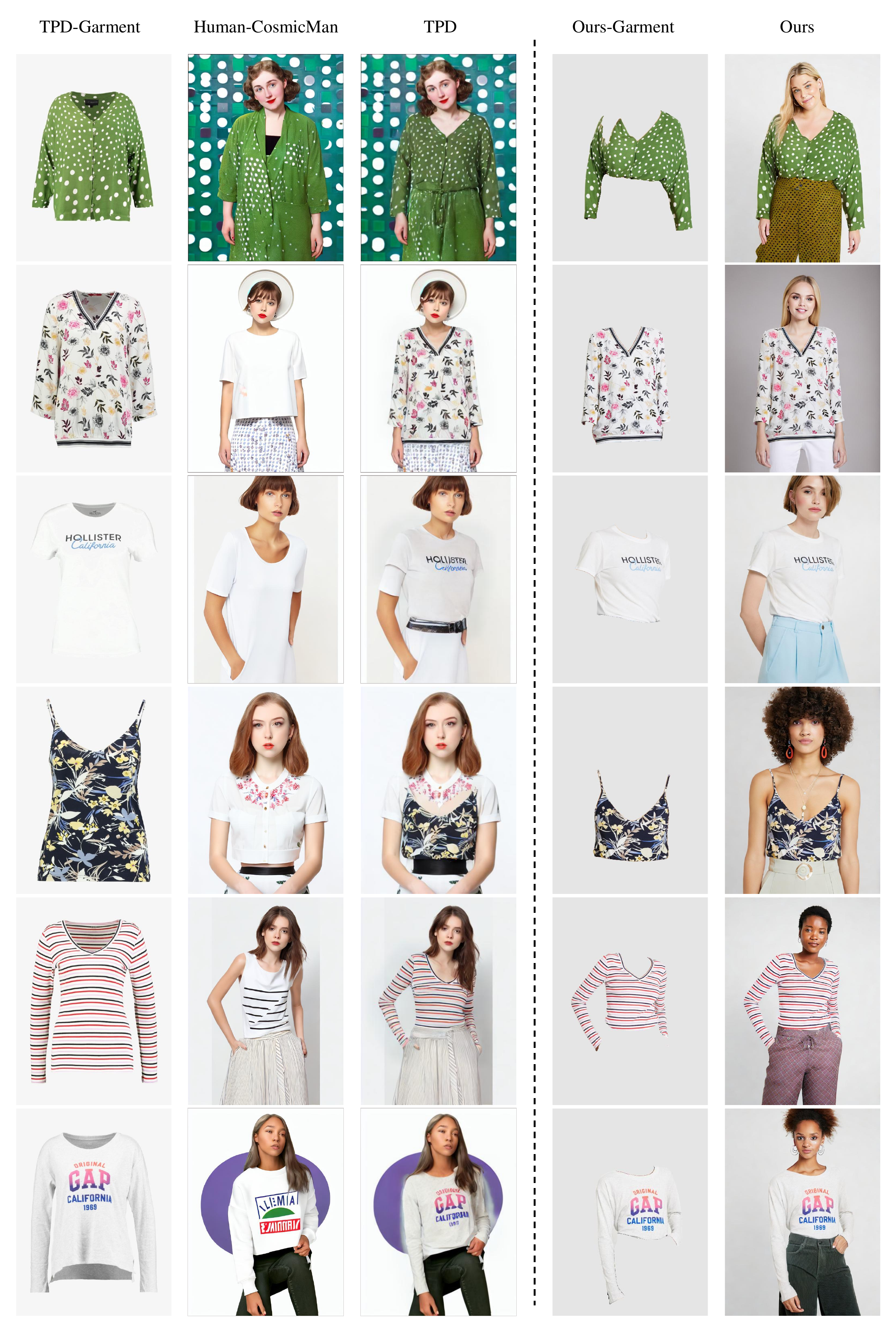}
    \caption{Qualitative comparison with CosmicMan+TPD.}
    \label{fig:tpd}
\end{figure*}

To evaluate the performance of our method and virtual try-on (VTON) based methods on the task of fashion image generation, we conduct an experiment with a two-stage VTON-based pipeline CosmicMan+TPD. In the first stage, we generate a virtual fashion model to show the target garment using a state-of-the-art text-to-image human generation framework CosmicMan~\cite{li2024cosmicman}. In the second stage, we utilize the VTON method TPD~\cite{yang2024texture} to generate the showcase image in which the above virtual fashion model wears the target garment. This pipeline mimics the showcase image generation capabilities of our framework. The results reveal that FashionMAC achieves outstanding generation performance, excelling in preserving human structural features and garment fidelity.~\figref{fig:tpd} demonstrates some generated showcase images with CosmicMan+TPD and our method, respectively.

\section{Fashion Image Generation for Mannequins}
\begin{figure*}[!h]
    \centering
    \includegraphics[width=\linewidth]{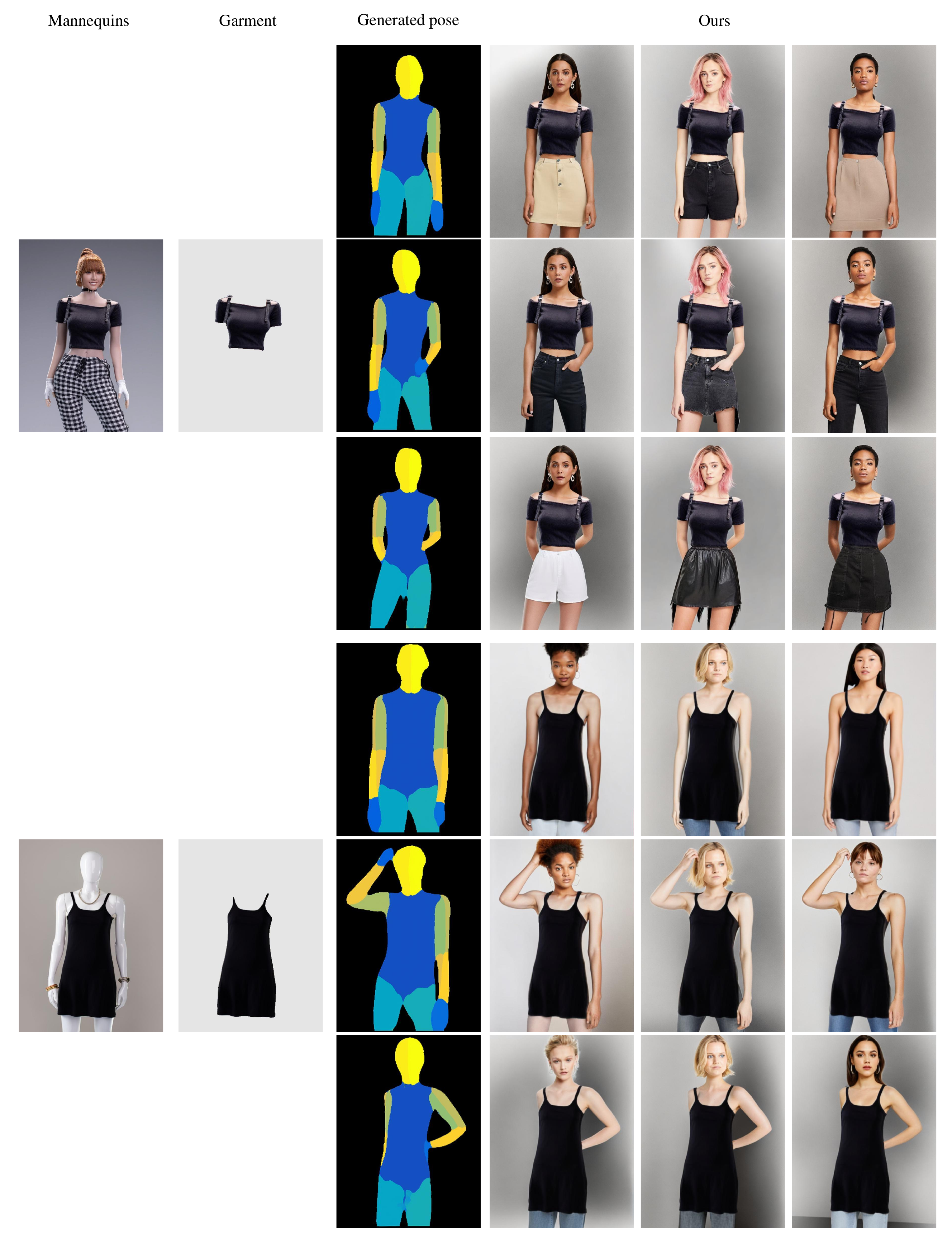}
    \caption{The results of our method conditioned on garments worn by a mannequin.}
    \label{fig:mannequin}
\end{figure*}

In e-commerce, the retailers usually use mannequins to coarsely demonstrate the garments. Our FashionMAC can utilize mannequin garment images as input and generate realistic apparel showcase images. For an input mannequin image that wears a garment, we can obtain the target garment through a pretrained segmentation network. Then FashionMAC can generate the corresponding pose map and produce high-quality fashion images.~\figref{fig:mannequin} shows the results conditioned on mannequin images. The results indicate the robustness and generalization ability of our method.


\section{Applicability of FashionMAC}

The goal of FashionMAC is to generate high-quality garment showcase images for e-commerce retailers. The key points of the task are (i) preserving garment details and (ii) controllable model appearance customization. Given a fixed segmented garment, the diversity of body shapes and the poses of the synthesized models are limited. Yet, FashionMAC can accept photos of users or plastic mannequins wearing the garment with diverse body shapes and poses as inputs, and then customize the model appearances from these inputs, which are easy to obtain in real e-commerce scenarios. Currently, FashionMAC's performance across different body shapes and poses is constrained by limited dataset diversity. However, as FashionMAC does not require paired training data as in VTON, expanding training data to include a wider range of body shapes should substantially improve its generalization.

\section{Limitation and Future Work.}

Since current fashion datasets contain data biases, such as the predominance of young female fashion models, the range of controllable attributes achievable by our method trained on these datasets is limited. Expanding these datasets to include a broader variety of ages, body types, genders, cloth types and backgrounds would be a promising direction for future work, as it would enhance the versatility and controllability of our method in generating more diverse and representative outputs.

\end{document}